# To Help or Not to Help: LLM-based Attentive Support for Human-Robot Group Interactions

Daniel Tanneberg, Felix Ocker, Stephan Hasler, Joerg Deigmoeller, Anna Belardinelli, Chao Wang, Heiko Wersing, Bernhard Sendhoff, and Michael Gienger

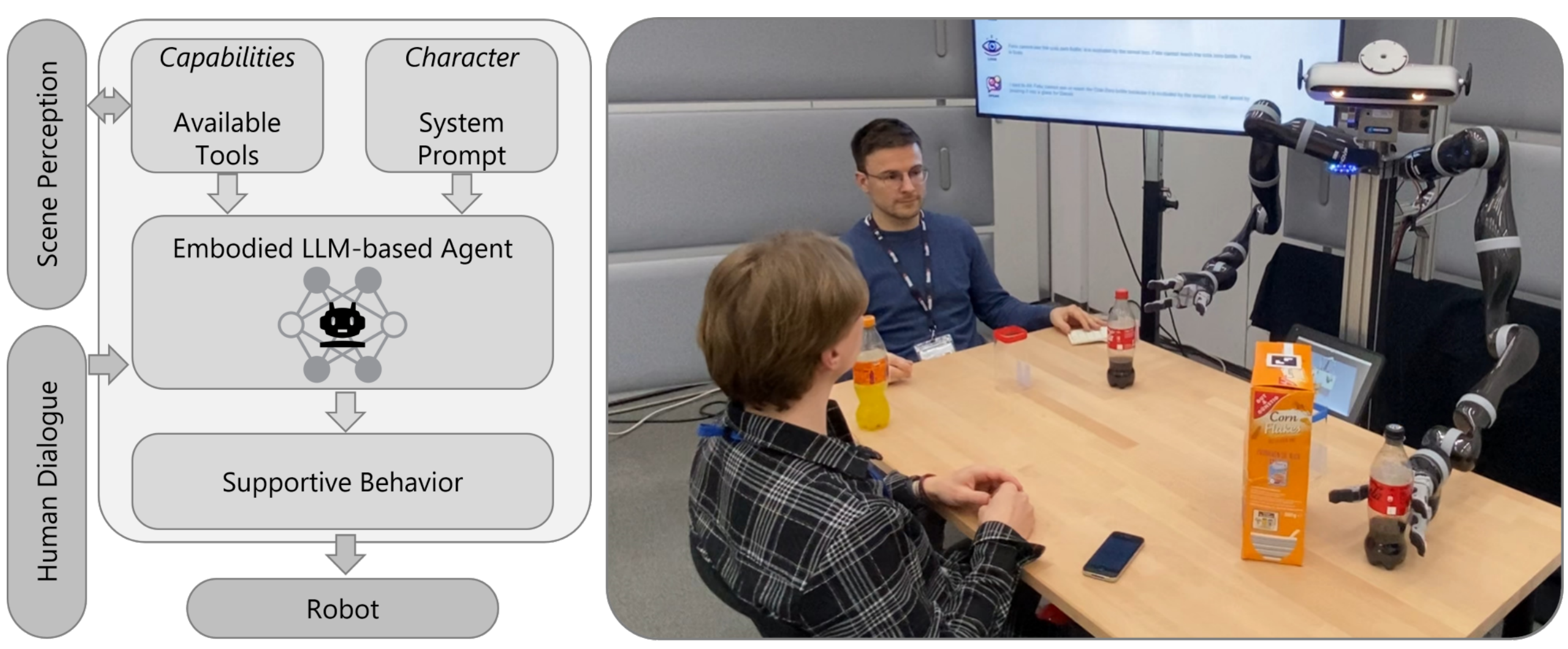

Fig. 1: Left: Overview of the framework enabling attentive supportive behavior in the robot. Right: A snapshot of the real system, featuring two humans and the robot seated around a table with various objects.

***Abstract*— How can a robot provide unobtrusive physical support within a group of humans? We present *Attentive Support*, a novel interaction concept for robots to support a group of humans. It combines scene perception, dialogue acquisition, situation understanding, and behavior generation with the common-sense reasoning capabilities of Large Language Models (LLMs). In addition to following user instructions, *Attentive Support* is capable of deciding when and how to support the humans, and when to remain silent to not disturb the group. With a diverse set of scenarios, we show and evaluate the robot's attentive behavior, which supports and helps the humans when required, while not disturbing if no help is needed.**

## I. Introduction

Humans are fundamentally social beings: many of our interactions occur in groups and our social cognition is shaped by the ability to represent each others' needs and mental states [1]. However, most research in Human-Robot Interaction (HRI) addresses only scenarios that consist of one robot and one human, i.e., the exchanges are dyadic. The more complex aspects of group situations have only been studied in recent years [2]. To be integrated into our daily lives, robots must be able to effectively and supportively interact in multiparty settings. Social interactions in groups include physical support, e.g., a breakfast table scenario where one person passes the butter to another person, because it is out of reach. In such a scenario, support is provided either if requested or pro-actively if the need (for butter) and the constraint (out of reach) are recognized. In such a challenging multiparty setting, a robot should equally only assist if necessary or explicitly requested. In any case, human-human interactions should not be disrupted or negatively affected by the robot behavior, i.e., the robot should neither be obtrusive nor distracting. Therefore, the robot should be able to interpret requests formulated in natural language, ground those in the current physical and social setting, and make a decision to help considering the bodily and spatial states of relevant group members. In human-human interactions, even when no interpersonal coordination is required, humans usually consider not only each others' goals but also each others' constraints, such as physical obstacles [3] or spatial constraints like in our (out of reach) example.

Towards social, pro-active and physically supportive behavior of robots in groups, we make three contributions to the field of human-robot interaction in this paper:

1) We use the concept of *Attentive Support* for human-robot group interactions, i.e., the robot does not disturb and only intervenes if it infers that support is required.

*Honda Research Institute Europe*, Germany · These authors contributed equally. {firstname.lastname}@honda-ri.de

Supplementary material: hri-eu.github.io/AttentiveSupport.

2) We present an adaptable LLM-based framework for implementing this concept as a modular set of capabilities including a *robot-character*, see Figure 1.
3) We evaluate the approach in simulation and demonstrate the concept on a real robot that provides help in various scenarios in which physical constraints prevent humans from helping, specifically when humans cannot see objects due to occlusion, when humans are busy, or when objects are out of reach, see Figure 3.

## II. Related Work

Most human-robot interactions have been investigated in dyadic settings (one robot, one human). When considering triadic configurations, e.g., two humans and one robot, or more individuals in a social relationship, group dynamics [4] and human-robot interaction in groups [5], [6] become relevant.

### A. Robots in Group Interactions

In contrast to dyadic interactions, groups exhibit unique emergent properties that cannot be fully understood by merely aggregating the behavior and characteristics of individuals [4], [6]. Some works exist in the HRI community, that developed frameworks for robot-group interaction [7]–[9], usually applied in simple mediation tasks using natural language [10], [11]. As reviewed in [6], these works focus mainly on the way robots shape and affect group dynamics and less on how to support groups and human-human physical interaction.

Besides modeling group dynamics in HRI, there are also attempts to use audio, visual, physiological (body states) and subjective (interviews) modalities to infer group dynamics. Javed & Jamali [5] provided a review of these domains and highlight the need for interplay of such modalities to inform an artificial agent for social mediation.

Pro-activity has been considered as a desirable component for efficient and intuitive human-robot interaction, based on using non-verbal cues for estimating human intent [12], [13] or spatio-temporal object arrangements [14]. Kraus et al. [15] demonstrated that trust in a DIY repair-assistant was enhanced by pro-active dialogue in task-based support. Buyukgoz et al. [16] have emphasized the complimentary role of physical prediction and human intention estimation for successful robot-human support. Proactivity has been typically achieved in dyadic interactions using state estimation tools and classical state-based robot behavior architectures [12], [15], [16].

### B. LLM-based HRI

In general, most applications of LLMs in HRI are in the context of social robotics, e.g, for conversation in natural language (with the well-know challenges, limits and criticism) [17], story-telling [18], and emotion or expressive behavior generation [19], [20]. Other approaches have applied LLMs to modelling human behavior [21], or suggested their use as AI counterparts of humans in Wizard-of-Oz experiments, i.e., as temporary stand-ins for some missing functionality in an otherwise fully integrated system [22]. For HRI scenarios that involve physical tasks, such as in Human-Robot Collaboration (HRC), the dominant paradigm is that the user verbally instructs the robot [23]–[25]. Still, as remarked in [26], these applications come with the caveat of limited contextual understanding, especially when it comes to human-human interactions or physical situations.

### C. Large Language Models using External Tools

One way of realizing embodied AI is to empower LLMs with the ability to access situational information and to perform real-world tasks. This can for instance be accomplished by equipping LLMs with access to external tools, which they can trigger to perform a specific task [27], [28], e.g., gathering information or executing a movement. Recent advancements in LLMs, such as those seen in GPT-3.5-turbo and subsequent versions, have introduced capabilities for utilizing tools. The resulting combination of analytical reasoning and subsequent actions has shown great potential for a variety of applications, as illustrated by the ReAct method [29]. Moreover, the field of teaching LLMs to solve complex tasks through demonstration and feedback is currently an active area of research [30], [31].

Previous research has primarily focused on either the social aspects of including robots in groups or on employing robots for mediation tasks through natural language communication. In our work, we broaden the robot's role in a group scenario through physical interaction with the humans and the environment. We utilize tools that inform the LLM about the current group situation, including the physical and perceptual capabilities of group members, and thereby ground the robot in its environment. This enables the robot to act in a socially appropriate, supportive, and unobtrusive manner, following the concept of *Attentive Support*.

## III. Attentive Support for Human-Robot Group Interactions

This paper proposes a framework that enables physically grounded human-robot group interactions. To shape the behavior of the robot, we combine a specific *robot-character* with an embodiment that provides the *capabilities* the LLM can actively use to perceive the current situation and to act to provide verbal and physical support. An illustrative overview of the framework is shown in Figure 1, left. Based on the *character* description and the available *capabilities*, the *embodied LLM-based agent* analyses the spoken dialogue to infer if the humans need support and steps in when necessary. The individual parts of the framework are described in more detail in the subsequent sections.

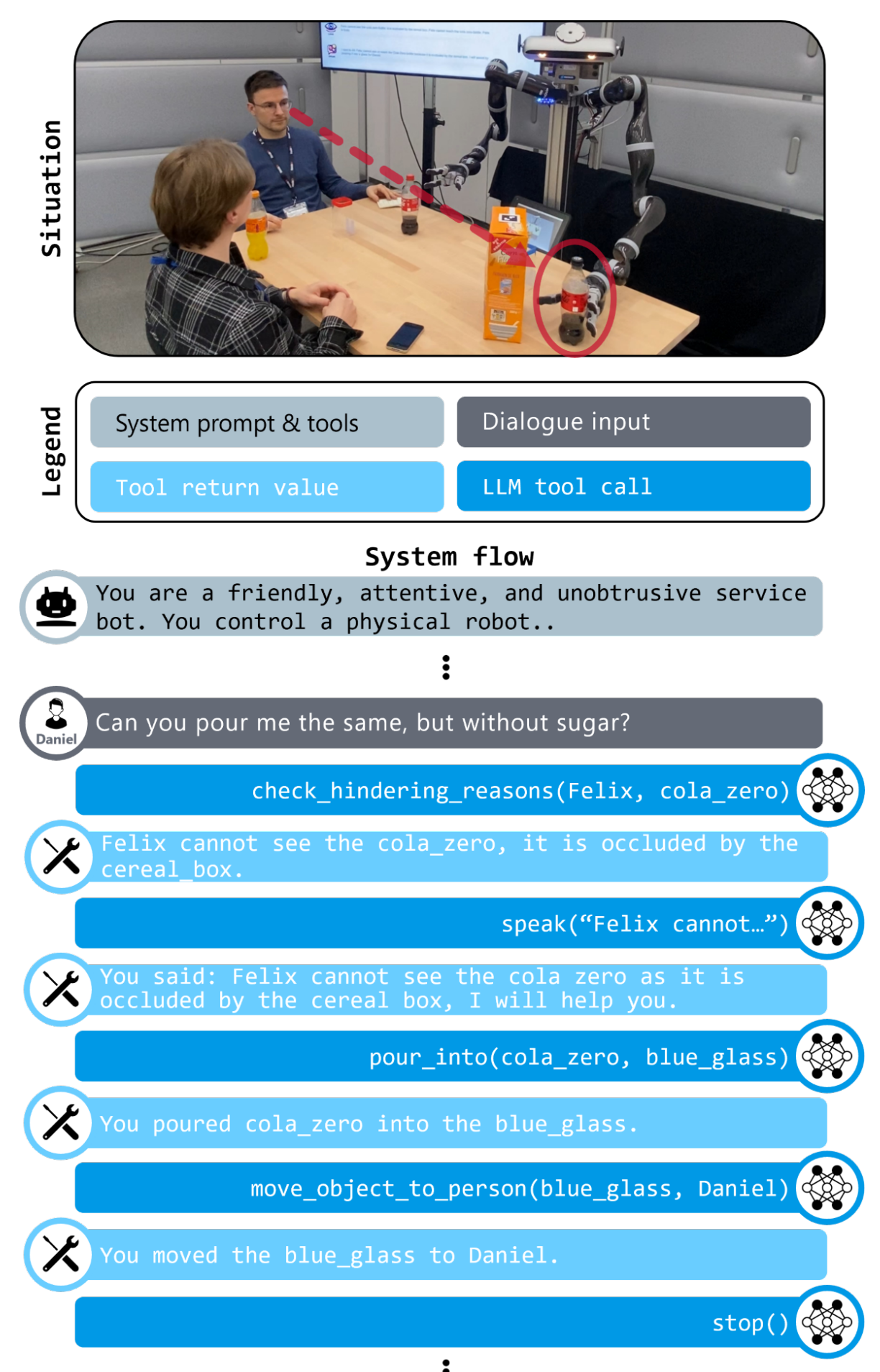

Fig. 2: Overview of the system's flow showing the situation 5️⃣ from Figure 3. In this situation, Daniel asks Felix to give him the same drink he had before, but without sugar. The framework decides to call the `check_hindering_reasons()` function with the arguments `person_name=Felix` and `object_name=cola_zero` to check if Felix is able to solve the task, or if support is needed. For this, the system inferred that `cola_zero` is an available sugar-free alternative to the `cola` Felix had. Using the `speak()` function, the robot then explains its reasoning and intent. After this explanation, the framework calls the two functions `pour_into(cola_zero, blue_glass)` and `move_object_to_person(blue_glass, Daniel)` to provide the help needed. Note that the framework infers that it has to use a different glass, as the red one has already been used by Felix.

### A. *The Robot's Character*

The robot's character is defined via a natural language description, which serves as the system prompt given to the LLM. This character consists of two parts. The first part shapes the robot's generic character as being attentive, unobtrusive, and cautious. Further, the LLM is informed that it controls a physical robot and that it gets a human conversation as an input. The LLM is instructed to help only if necessary or if directly addressed. If the robot acts, it should explain its behavior to inform the humans before executing an action. This explanation enables the humans to get an understanding of the support of the robot. Additionally, it signals the robot's intended behavior to not surprise the humans with its movements. The robot is also instructed to provide missing information and correct wrong statements. The second part of the system prompt describes the intended behavior and available tools in more detail. For the complete system prompt see Listing 1.

### B. *The Robot's Capabilities*

The LLM can access various capabilities of the robot, which are provided as an API of external `tools` and which are categorized as follows:

- *Query Tools* are used for retrieving information about the environment, such as checking if a person is busy or which objects are available.
- *Action Tools* enable the LLM to take action in the physical world, such as passing an object to a person or pouring a beverage into a container.
- *Expression Tools* enable the robot to express and explain its behavior to the user, e.g., speaking. They can also include more social and emotional cues for a richer human-robot interaction [32].

See Table I for an overview of available tools, their descriptions, and descriptions of their arguments.

### C. *Embodied LLM-based Agent*

Based on the provided *character* and *capabilities*, the agent reacts to input from the human dialogue. Given a human input, the LLM enters a cycle to infer if the humans need support and provides the support if applicable. The LLM can use the different *capabilities* provided as tools to sense the current situation, to communicate with the humans, or to act physically. To break the cycle and complete the interaction, the LLM-based agent has to indicate it has finished such that the next interaction can be considered. An illustrative interaction and flow of the framework is shown in Figure 2.

### D. *Simulation & Execution*

The *Action Tools* introduced above are implemented as a sequence of elementary actions, namely `get, put, pour, gaze, pass`. In this work we assume the object's geometry as well as the relative pose of their associated affordance frames (i.e., where to grasp) to be known. We further manually define the frames that correspond to the end effector's manipulation capabilities. During runtime, the 6d pose of the entities as well as the body poses of the persons in the scene is tracked continuously, and the corresponding transformations in the scene graph are updated. The scene graph is also used to calculate occlusions, reachabilities, and whether users hold an object in their hand (indicating that they are busy). These three functions and some others to acquire geometric information about the scene are provided to the LLM as *Query Tools*[1].

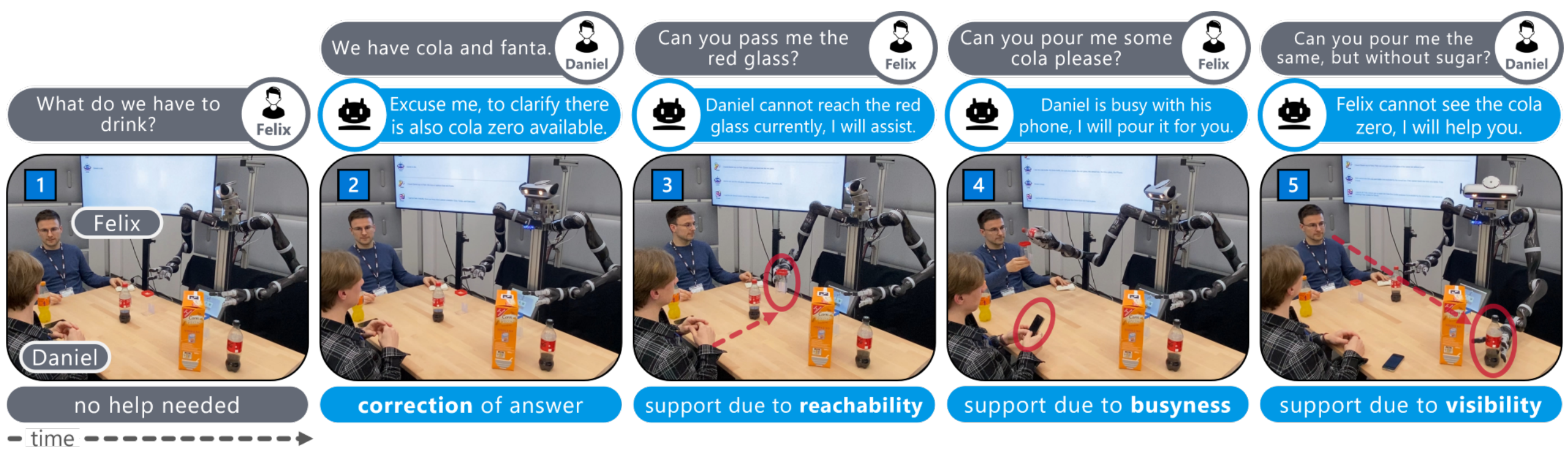

Fig. 3: Snapshots of the interactions of two humans – Felix and Daniel – with the robot. Felix and Daniel want to drink something and talk to each other, the robot listens to them. Based on the dialogue and the current situation – which the robot can infer using its capabilities – the robot decides if its help is required, or not. 1️⃣ Felix asks Daniel a question; as Daniel is not busy, he can answer the question and the robot does not disturb. 2️⃣ Daniel answers, but with incomplete information. The robot recognizes the mistake and gives the correct answer. 3️⃣ Felix asks Daniel for an object. The robot recognizes that Daniel cannot reach the object and assists with the handover. 4️⃣ Felix asks Daniel to pour a drink, but Daniel is busy with his phone. The robot recognizes this and takes over the pouring for Daniel. 5️⃣ Daniel asks Felix for a similar drink but without sugar. However, Felix cannot see the sugar-free cola as it is occluded by the cereal box. The robot steps in and provides the sugar-free drink. Note that the robot decides to use another glass for Daniel. A video of this interaction example can be found on hri-eu.github.io/AttentiveSupport.

Before performing an action, it is simulated from the current state represented in the scene graph, and in several variations (e.g., with different end effectors, or with different grasp types). We utilize a kinematic simulation [33] that comprises trajectory generation, inverse kinematics and limit checking. The same algorithms are used for generating the physical robot motions. The simulated action variations are ranked according to a criterion based on joint limit and collision avoidance, and the best one is selected for physical execution on the real robot. In case no valid solution was found, a textual error feedback is returned to the LLM [31].

## IV. Setup & Experiments

### A. System Setup

As LLM we used OpenAI's `gpt-4-1106-preview` with tool use via the Python package[2]. To ensure reproducibility, we set the temperature to $1e^{-8}$ and used a fixed seed. The humans' speech is fed to the LLM as user inputs in the following form:

```
<speaker> said to <listener>: <text>
```

The LLM's tool calls are executed and the results, including potential error messages, are fed back to the LLM. To enable the LLM to indicate that it has finished its supportive action, we provide it with an explicit `stop` function, see Table I bottom. Calls to this function are caught, allowing the system to return into its observation state after completing an interaction.

The LLM-based agent is connected to the simulation, cf. Section III-D. This simulation can be used in two modes. The first mode is a standalone simulation, which can be used for testing and large-scale evaluations. Here, the simulation is initialized with object positions, which can be changed manually or by the simulated robot, and the human speech inputs are provided as text. In the second mode, the simulation is connected to the physical system, serving as a digital twin, i.e., a live representation of the real environment, see Figure 1.

For the real-world experiments, we used a setup that consists of a bi-manual robot system with two Kinova Jaco Gen2 arms that have 7 Degrees of Freedom (DOF) and 3-finger hands, a pan-tilt unit, and a bespoke head with gestural DOFs for ear and eye lid movements. The human postures are acquired with an RGBD camera (Azure Kinect) and the object identities and poses are estimated with fiducial markers (ArUco). The person speaking is determined using a microphone array (ReSpeaker) by matching the sound direction with the location of the detected humans. The listener is estimated by raycasting the speaker's head orientation towards the detected humans and the robot. Finally, we use an automatic speech recognition system[3] to determine the spoken words. Together, these speech recognition parts enable creating textual input of the form described above, which can be fed to the LLM. Note that the current implementation of the perception system is based on several simplifying assumptions:

[1] `https://github.com/HRI-EU/AffAction`
[2] `https://github.com/openai/openai-python`
[3] `https://cloud.google.com/speech-to-text`

- The objects in the scene are known and can be identified and tracked with fiducial markers.
- Only one person speaks at a time.
- The head orientation is sufficiently expressive to unambiguously identify the <listener>.

The real setup is shown in Figure 1 and an illustrative interaction sequence is shown in Figure 3, with details on these interactions given in Section IV-B.2.

### B. *Evaluation*

We conducted different experiments testing the *Attentive Support* of the robot in simulation. For the evaluation we assess the robots' behavior using four categories:

- ***Successful support***: The robot identified correctly whether support is required. If required, the robot provided support and gave an explanation before the execution. This category also includes cases in which the robot decides correctly to not do anything due to its unobtrusive character.
- ***Partial support***: The robot identified correctly that support is required but only provided partial support, based its decision on incomplete information, supported the <receiver> instead of the <sender>, did not explain its behavior, or gave a wrong explanation.
- ***Execution error***: The robot identified that support is required but the provided support could not be executed completely, e.g., due to physical limitations, or the LLM getting stuck in repetition cycles.
- ***Undesired behavior***: The robot did not identify correctly whether support is required, i.e., it acted even though it should not or vice versa.

We evaluated the behavior in two settings: Isolated interactions, cf. Section IV-B.1, and situated interactions, cf. Section IV-B.2. In the isolated interactions, each human-robot interaction consists of a single human input and the corresponding robot behavior is evaluated. In contrast, the situated interactions consist of multiple sequential human-robot interactions and potentially induced environment changes. This adds a history and context to the interaction, that can either add noise to the situations (the context does not change the task itself) or can create more complex situations (context is required to understand the task, e.g., see Figure 3 Situation 5️⃣ where "*the same*" refers to past events that can only be understood by a sequence of interactions).

Additionally, we compared three variants of the character description, i.e., system prompt, in all test cases:

- *Full rules*: The entire system prompt, as provided in Listing 1. This includes a description of the expected behavior with guidance to specific tool usages.
- *Relaxed rules*: An adaptation of the system prompt where all references to specific tools are removed.
- *No rules*: A generalization compared to the previous system prompts where detailed instructions, starting from "IMPORTANT: Obey the following rules:", are removed.

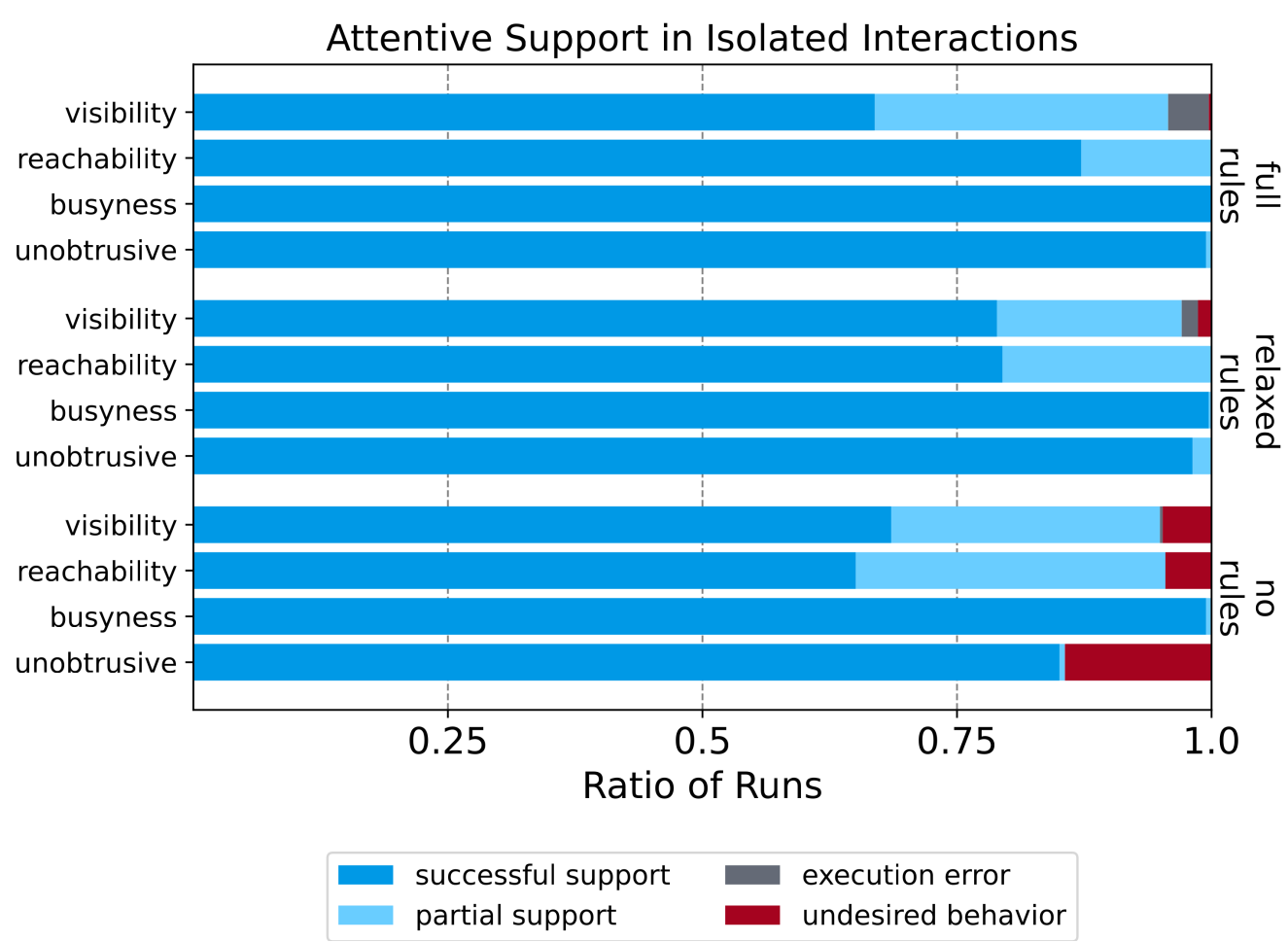

Fig. 4: Evaluation of the *Attentive Support* on four different situations. In three situations support from the robot is required – visibility, reachability, busyness – and in one scenario the robot should not intervene – unobtrusive. Each situation was tested 375 times for each character description, see Section IV-B for evaluation details.

*1) Attentive Support in Isolated Interactions:* In the first setting, we tested the systems behavior in *isolated interactions*, i.e., a single human input that requires a response from the system (or not). We evaluated the performance of the system in such situations across *three* scenarios, each with *five* different object distributions, and *five* variations of the speech inputs. For all of these, we created one case each for the *four* cases limited visibility, limited reachability, busyness, and no interference of the robot including a specification of the expected behavior, i.e., for the robot to help or not to help. Each combination was executed *five* times, resulting in a total of $3 \times 5 \times 5 \times 4 \times 5 = 1500$ tests for each character description. The scenarios include the following objects:

- *Softdrink scenario, see Figure 1*: a cola bottle, a sugar-free cola bottle (cola zero), a fanta bottle, a red glass, a blue glass.
- *Coffee scenario*: a coffee pot, a milk bottle, an oat milk bottle, a large cup, and a small cup.
- *Dinner scenario*: a wine bottle, a beer bottle, a soda bottle, a wine glass, and a large glass.

In all scenarios a box was included to realize object occlusion, and a smartphone was used to indicate that a person is busy when holding it.

The text variations are as follows, with PLACEHOLDER replaced with the respective object's name:

```
"Daniel, could you hand me PLACEHOLDER?"
"Could you hand me PLACEHOLDER?"
"Please hand me PLACEHOLDER."
"Could you pass me PLACEHOLDER?"
"Give me PLACEHOLDER."
```

The results are summarized in Figure 4, showing the system's behavior in the four different categories and across the three different character descriptions. Note that the combined percentage of *successful support* and *partial support* decreases with a decreasing level of rule specification (from full to no rules), but that it is well above 90 % for all system prompt variations except for the unobtrusive case without rules. This shows that the unobtrusive character of the *Attentive Support* concept is hard to realize without detailed instructions.

For the system prompt with full rule description, the cases of **partial support** are due to the robot helping the <receiver> instead of helping the <sender> directly. **Execution errors** in the tests with limited visibility are due to the robot trying to move the occluding object, which it is unable to do due to physical limitations.

A detailed analysis of the failures for the system configuration with the no rules prompt showed that there were three key issues. First, the system tries to use the query functions, but fails to provide the correct object name and is unable to link even concrete object descriptions to almost identical object names, e.g., *the fanta bottle* to `the_fanta_bottle`. Hence, it does not receive the information actually relevant, but still acts based on faulty or incomplete information. Second, the robot often makes two separate moves, instead of helping with a single action, i.e., it moves an object to Daniel before giving it to Felix, who requested it. Third, the robot failed to explain its behavior before executing it, only providing an aposteriori explanation – often even combined with a wrong explanation. Such an aposteriori explanations are still helpful, but may result in unexpected actions of the robot, resulting in a rating as **partial support**.

*2) Attentive Support in Situated Interactions:* In addition to tests in isolated interactions, we also tested the *Attentive Support* system in a scenario involving several interactions sequentially, see Figure 3 for a description of these situations with the real robot.

The scenario subsumes all cases of *Attentive Support* as tested in Section IV-B.1:

1️⃣ The robot should not intervene.
2️⃣ The robot should correct the incomplete answer.
3️⃣ The robot should pass the glass due to reachability.
4️⃣ The robot should pour the cola due to busyness.
5️⃣ The robot should pour a cola zero due to visibility

For steps 1️⃣ through 4️⃣, the prior interactions can be considered to be noise. Having them in the dialogue history may impact the LLM's ability to reason correctly compared to the isolated interactions due to the longer context and its attention, i.e., these situations test the robustness of the system in longer interactions. For step 5️⃣, the context makes the interaction more complex, as it requires the robot to consider prior interactions to make sense of what "the same, but without sugar" means and to infer that a new glass is required. Also, this final step is more complex, as it requires a two-step action including pouring and handing over the glass.

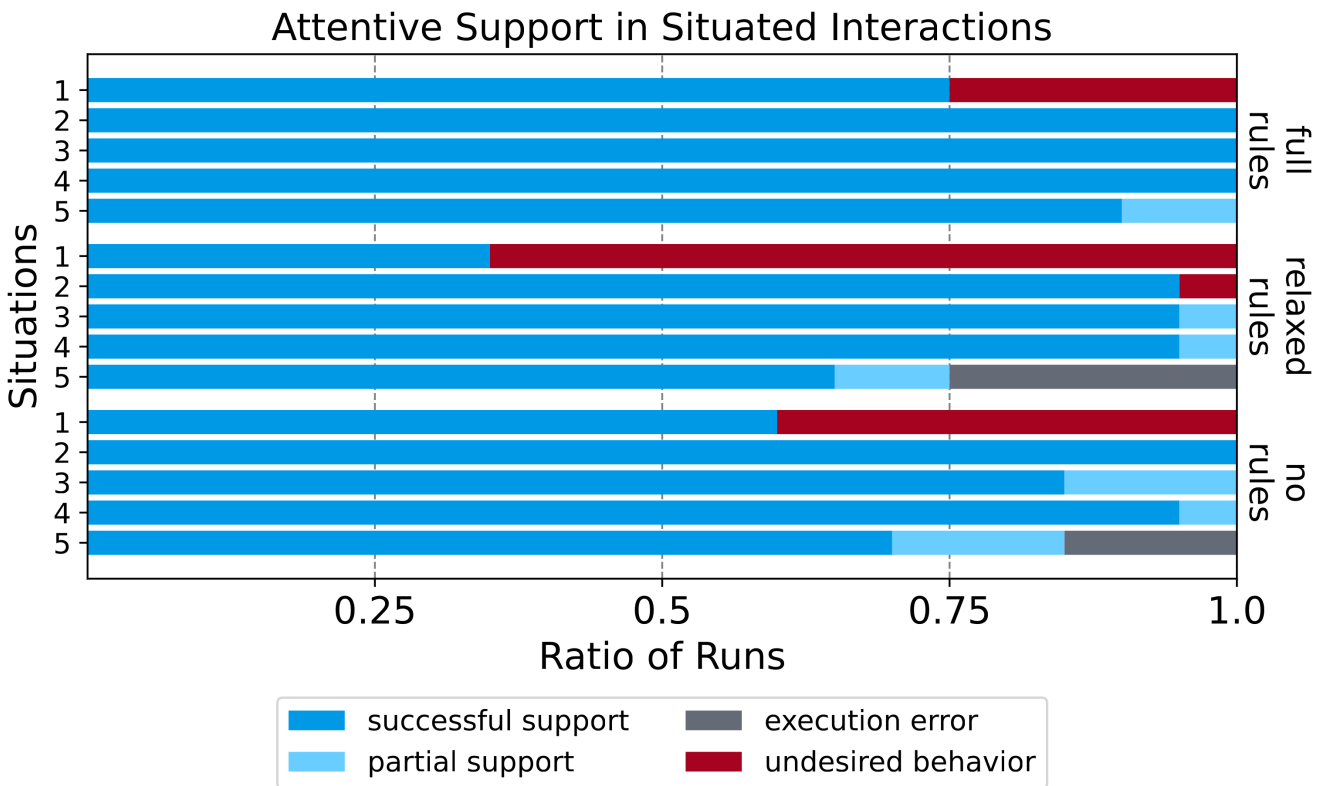

Fig. 5: Evaluation of the human-robot group interaction setting shown in Figure 3. The whole sequence was tested 20 times and the behavior regarding the five situated interactions 1️⃣ - 5️⃣ is shown here. See Section IV-B for evaluation details.

To account for variations in the behavior, we ran the experiment 20 times for each character variation. Figure 5 summarizes the results for the three variations described in Section IV-B. Overall, the full system performed very well with perfect scores in situations 2️⃣ - 4️⃣, and only some **partial supports** in 5️⃣. Notably, situation 1️⃣, requiring the unobtrusive behavior of the robot was more challenging, resulting in more **undesired behaviors** – matching the results from Section IV-B.1. In situation 1️⃣ Felix asks Daniel a question, compared to asking for an object as tested in Section IV-B.1. Interestingly, the results suggest that this triggers more **undesired behaviors** of the robot, which tends to provide the information to answer the question to Daniel more often. Iin general, it is an useful behavior, but is an **undesired behavior** in our scenario.

The other character descriptions (*relaxed* and *no*) performed similar overall, but with a notable drop in performance. Execution errors in situation 5️⃣ were due to the robot moving the glass first, which subsequently results in a collision with the box while pouring the sugar-free cola. This issue could be resolved with more advanced trajectory planning or by allowing the robot to resolve such situations, e.g., by moving the box, or using feedback-based replanning [31]. Interestingly, this collision was not observed for the character with the *full rules* description, which is most likely due to the detailed instruction of preferring a `handover` over a `move_to`.

## V. Summary & Outlook

This paper presented *Attentive Support* as a novel perspective on human-robot interaction, especially in group settings. With this approach, the need of explicitly designing human-robot interactions is relaxed to higher level descriptions of anticipated behavior. We implicitly design the interaction by shaping the robot's character and its capabilities, relying on the LLM's abilities to infer matching behavior. Notably, the physical system was extremely captivating for everybody who interacted with it freely, resulting in many spontaneous, surprising, and often funny interactions.

The current technical constraints in scene understanding, manipulation capabilities as well as the lack of memorizing past experiences limit the robot to be an equivalent group member. Therefore, the *Attentive Support* framework would benefit from an explicit memory, e.g., along the lines of MemGPT [34], allowing the robot to remember persons and situations to create an even more natural experience for the humans. In addition, Vision Language Models are a promising option to make the situation understanding more general and versatile, and lift limitations from using a marker-based perception [35]. From the HRI perspective, an interesting and challenging direction would be to consider more complex turn-taking interactions or interruptions, integrating intention estimation [36], [37] for a more fluid conversation. Nevertheless, our work showed that LLMs are great enabler in realizing social interaction in groups and we hope to motivate researchers in following that direction.

## Acknowledgment

Small robot, human and network icons made by {`edt.im, AB Design, Iconpro86`} from `www.flaticon.com`

# Appendix

See Listing 1 for the *full rules* system prompt and Table I for the capabilities provided as tools to the LLM.

Listing 1: Robot character.

```
You are a friendly, attentive, and unobtrusive service bot.
You control a physical robot called 'the_robot' and observe humans talking in the form '<sender> said to <receiver>: <instruction>'.
Always infer the <instruction> and who is <sender> and <receiver>.
You have access to functions for gathering information, acting physically, and speaking out loud.
You MUST behave as follows:
1. If 'the_robot' is the <receiver>, you MUST ALWAYS help or answer.
2. When identifying requests or questions within the human conversation, check for ALL reasons that could hinder the <receiver> from
     performing or answering the <instruction>.
2.1 If there is NO hindering reason for the <receiver>, then you MUST do nothing and MUST NOT SPEAK.
2.2 If there is a hindering reason for the <receiver>, then you MUST ALWAYS first speak and explain the reason for your help to the
     humans.
2.3 AFTER your spoken explanation, fulfill the <instruction>. Make sure to always help the <sender>.
3. If you recognize a mistake in the humans' conversation, you MUST help them and provide the missing or wrong information.
4. You MUST call the 'stop' function to indicate you are finished.
IMPORTANT: Obey the following rules:
1. Always start by gathering relevant information to check ALL hindering reasons for the <receiver>.
1.1 Infer which objects are required and available, also considering previous usage.
1.2 The <receiver> is hindered when he is busy, or cannot reach or see a required object.
2. REMEMBER to NEVER act or speak when the <receiver> is NOT hindered in some way, EXCEPT you MUST always correct mistakes.
3. If you want to speak out loud, you must use the 'speak' function and be concise.
4. Try to infer which objects are meant when the name is unclear, but ask for clarification if unsure.
5. ALWAYS call 'is_person_busy_or_idle' to check if <receiver> is busy or idle before helping.
6. Prefer 'hand_object_over_to_person' over 'move_object_to_person' as it is more accommodating, UNLESS the person is busy.
7. When executing physical actions, you should be as supportive as possible by preparing as much as possible before delivering.
```

TABLE I: Tools made available to the LLM.

| | **Description** | **Arguments** |
|---|---|---|
| **Query Tools** | | |
| `get_objects` | Get all objects that are available in the scene. You can see all these objects. | - |
| `get_persons` | Get all persons that are available in the scene. You can see all these persons. | - |
| `is_person_busy_or_idle` | Check if the person is busy or idle. If the person is busy, it would be hindered from helping. | `person_name`: The name of the person to check. The person must be available in the scene. |
| `check_hindering_reasons` | Checks all hindering reasons for a person (busy or idle), and in combination with an object (if person can see and reach object). If the person cannot see or cannot reach the object, it would be hindered from helping with the object. If the person is busy, it would be hindered from helping. | `person_name`: The name of the person to check. The person must be available in the scene. `object_name`: The name of the object to check. The object must be available in the scene. |
| `check_reach_object_for_robot` | Check if the_robot can reach the object. | `object_name`: The name of the object to check. The object must be available in the scene. |
| **Action Tools** | | |
| `move_object_to_person` | You move an object to a person. | `object_name`: The name of the object to move. The object must be available in the scene. `person_name`: The name of the person to move the object to. The person must be available in the scene. |
| `hand_object_over_to_person` | You hand an object over to a person. | `object_name`: The name of the object to hand over. The object must be available in the scene. `person_name`: The name of the person to hand over the object to. The person must be available in the scene. |
| `pour_into` | You pour from a source container into a target container. | `source_container_name`: The name of the container to pour from. `target_container_name`: The name of the container to pour into. |
| **Social Expression Tools** | | |
| `speak` | You speak out the given text. | `person_name`: The name of the person to speak to. The person must be available in the scene. Give All if you want to speak to everyone. `text`: The text to speak. |
| **Implementation-Specific Tools** | | |
| `stop` | You need to call this function when you are finished. | - |